\documentclass[11pt,a4paper]{article}
\usepackage[hyperref]{emnlp2020}
\usepackage{times}
\usepackage{latexsym}
\usepackage{enumitem}
\usepackage{arydshln}
\usepackage{url}
\usepackage{multirow}
\usepackage{graphicx}
\usepackage{amsmath}
\usepackage{amssymb}
\usepackage{listings}
\usepackage{microtype}

\aclfinalcopy 
\def\aclpaperid{58} 

\title{Wikipedia2Vec: An Efficient Toolkit for Learning and Visualizing\\ the Embeddings of Words and Entities from Wikipedia}
\author{
    Ikuya Yamada$^{1,2}$\\
    {\small \texttt{ikuya@ousia.jp}}
    \And
    Akari Asai$^3$\\
    {\small \texttt{akari@cs.washington.edu}}
    \And
    Jin Sakuma$^4$\\
    {\small \texttt{jsakuma@tkl.iis.u-tokyo.ac.jp}}
    \AND
    Hiroyuki Shindo$^{5,2}$\\
    {\small \texttt{shindo@is.naist.jp}}
    \And
    Hideaki Takeda$^6$\\
    {\small \texttt{takeda@nii.ac.jp}}
    \And
    Yoshiyasu Takefuji$^7$\\
    {\small \texttt{takefuji@sfc.keio.ac.jp}}
    \And
    Yuji Matsumoto$^{2}$\\
    {\small \texttt{matsu@is.naist.jp}}
    \AND
    \begin{minipage}{\textwidth}
        \begin{center}
            \fontsize{11.5}{14}\selectfont
            \textnormal{$^1$Studio Ousia\,\,\,$^2$RIKEN AIP\,\,\,$^3$University of Washington\,\,\,$^4$The University of Tokyo}\\
            \textnormal{$^5$Nara Institute of Science and Technology\,\,\,$^6$National Institute of Informatics\,\,\,$^7$Keio University} \end{center}
    \end{minipage}
}

\date{}

\begin{document}
\maketitle
\begin{abstract}
    The embeddings of entities in a large knowledge base (e.g., Wikipedia) are highly beneficial for solving various natural language tasks that involve real world knowledge.
    In this paper, we present Wikipedia2Vec, a Python-based open-source tool for learning the embeddings of words and entities from Wikipedia.
    The proposed tool enables users to learn the embeddings efficiently by issuing a single command with a Wikipedia dump file as an argument.
    We also introduce a web-based demonstration of our tool that allows users to visualize and explore the learned embeddings.
    In our experiments, our tool achieved a state-of-the-art result on the KORE entity relatedness dataset, and competitive results on various standard benchmark datasets.
    Furthermore, our tool has been used as a key component in various recent studies.
    We publicize the source code, demonstration, and the pretrained embeddings for 12 languages at \url{https://wikipedia2vec.github.io}.
\end{abstract}

\section{Introduction}

Entity embeddings, i.e., vector representations of entities in knowledge base (KB), have played a vital role in many recent models in natural language processing (NLP).
These embeddings provide rich information (or \textit{knowledge}) regarding entities available in KB using fixed continuous vectors.
They have been shown to be beneficial not only for tasks directly related to entities (e.g., entity linking \cite{Yamada2016,ganea-hofmann:2017:EMNLP2017}) but also for general NLP tasks (e.g., text classification \cite{Yamada2019NeuralClassification}, question answering \cite{Poerner2019BERTQA}).
Notably, recent studies have also shown that these embeddings can be used to enhance the performance of state-of-the-art contextualized word embeddings (i.e., BERT \cite{devlin2018bert}) on downstream tasks \cite{Zhang2019ERNIE:Entities,Peters2019KnowledgeRepresentations,Poerner2019BERTQA}.

In this work, we present \textit{Wikipedia2Vec}, a Python-based open source tool for learning the embeddings of words and entities easily and efficiently from Wikipedia.
Due to its scale, availability in a variety of languages, and constantly evolving nature, Wikipedia is commonly used as a KB to learn entity embeddings.
Our proposed tool jointly learns the embeddings of words and entities, and places semantically similar words and entities close to one another in the vector space.
In particular, our tool implements the word-based skip-gram model \cite{Mikolov2013,Mikolov2013a} to learn word embeddings, and its extensions proposed in \newcite{Yamada2016} to learn entity embeddings.
Wikipedia2Vec enables users to train embeddings by simply running a single command with a Wikipedia dump file as an input.
We highly optimized our implementation, which makes our implementation of the skip-gram model faster than the well-established implementation available in gensim \cite{rehurek_lrecb} and fastText \cite{TACL999}.

Experimental results demonstrated that our tool achieved enhanced quality compared to the existing tools on several standard benchmarks.
Notably, our tool achieved a state-of-the-art result on the entity relatedness task based on the KORE dataset.
Due to its effectiveness and efficiency, our tool has been successfully used in various downstream NLP tasks, including entity linking \cite{Yamada2016,eshel-EtAl:2017:CoNLL,Chen2019YELM:Linking}, named entity recognition \cite{sato-EtAl:2017:I17-2,Lara-Clares2019}, question answering \cite{10.1007/978-3-319-94042-7_10,Poerner2019BERTQA}, knowledge graph completion \cite{Shah2019AnModels},  paraphrase detection \cite{Duong2019ADetection}, fake news detection \cite{Singh2019OnArticles}, and text classification \cite{Yamada2019NeuralClassification}.

We also introduce a web-based demonstration of our tool that visualizes the embeddings by plotting them onto a two- or three-dimensional space using dimensionality reduction algorithms.
The demonstration also allows users to explore the embeddings by querying similar words and entities.

The source code has been tested on Linux, Windows, and macOS, and released under the Apache License 2.0.
We also release the pretrained embeddings for 12 languages (i.e., English, Arabic, Chinese, Dutch, French, German, Italian, Japanese, Polish, Portuguese, Russian, and Spanish).

The main contributions of this paper are summarized as follows:
\begin{itemize}[leftmargin=10pt,topsep=-0pt,itemsep=-5pt]
    \item We present Wikipedia2Vec, a tool for learning the embeddings of words and entities easily and efficiently from Wikipedia.
    \item Our tool achieved a state-of-the-art result on the KORE entity relatedness dataset, and performed competitively on the various benchmark datasets.
    \item We present a web-based demonstration that allows users to explore the learned embeddings.
    \item We publicize the code, demonstration, and the pretrained embeddings for 12 languages at \url{https://wikipedia2vec.github.io}.
\end{itemize}

\section{Related Work}

Many studies have recently proposed methods to learn entity embeddings from a KB \cite{hu-EtAl:2015:ACL-IJCNLP,Li2016,tsai-roth-2016-cross,Yamada2016,TACL1065,C18-1016,cao-EtAl:2017:Long1,ganea-hofmann:2017:EMNLP2017}.
These embeddings are typically based on conventional word embedding models (e.g., skip-gram \cite{Mikolov2013}) trained with data retrieved from a KB.
For example, \newcite{ristoski2018rdf2vec} proposed RDF2Vec, which learns entity embeddings using the skip-gram model with inputs generated by random walks over the large knowledge graphs such as Wikidata and DBpedia.
Furthermore, a simple method that has been widely used in various studies \cite{yaghoobzadeh-schutze:2015:EMNLP,TACL1065,C18-1016,al2017tinkerbell,MasatoshiSUZUKI20182017SWP0005} trains entity embeddings by replacing the entity annotations in an input corpus with the unique identifier of their referent entities, and feeding the corpus into a word embedding model (e.g., skip-gram).
Two open-source tools, namely Wiki2Vec\footnote{\url{https://github.com/idio/wiki2vec}} and Wikipedia Entity Vectors,\footnote{\url{https://github.com/singletongue/WikiEntVec}} have implemented this method.
Our proposed tool is based on \newcite{Yamada2016}, which extends this idea by using neighboring entities connected by internal hyperlinks of Wikipedia as additional contexts to train the model.
Note that we used the RDF2Vec and Wiki2Vec as baselines in our experiments, and achieved enhanced empirical performance over these tools on the KORE dataset.
Additionally, there have been various relational embedding models proposed \cite{Bordes2013,wang-EtAl:2014:EMNLP20145,AAAI159571} that aim to learn the entity representations that are particularly effective for knowledge graph completion tasks.

\section{Overview}

\begin{figure}
    \centering
    \begin{lstlisting}[basicstyle=\ttfamily\scriptsize, frame=single, breaklines=true, language=bash]
$ wget https://dumps.wikimedia.org/enwiki/latest/enwiki-latest-pages-articles.xml.bz2
$ wikipedia2vec train enwiki-latest-pages-articles.xml.bz2 MODEL_FILE
\end{lstlisting}
    \caption{Shell commands to train embeddings from the latest English Wikipedia dump.}
    \label{fig:training}
\end{figure}

\begin{figure}
    \centering
    \begin{lstlisting}[basicstyle=\ttfamily\scriptsize, frame=single, breaklines=true, language=python]
>>> from wikipedia2vec import Wikipedia2Vec
>>> model = Wikipedia2Vec.load(MODEL_FILE)
>>> model.get_entity_vector("Scarlett Johansson")
memmap([-0.1979,  0.3086, ..., ], dtype=float32)
>>> model.get_word_vector("tokyo")
memmap([ 0.0161, -0.0332, ..., ], dtype=float32)
>>> model.most_similar(model.get_entity("Python (programming language)"))[:3]
[(<Word python>, 0.7265),
 (<Entity Ruby (programming language)>, 0.6856),
 (<Entity Perl>, 0.6794)]
\end{lstlisting}
    \caption{An example that uses the Wikipedia2Vec embeddings on a Python interactive shell.}
    \label{fig:api}
\end{figure}

\begin{figure*}[t]
    \centering
    \includegraphics[width=\textwidth,clip]{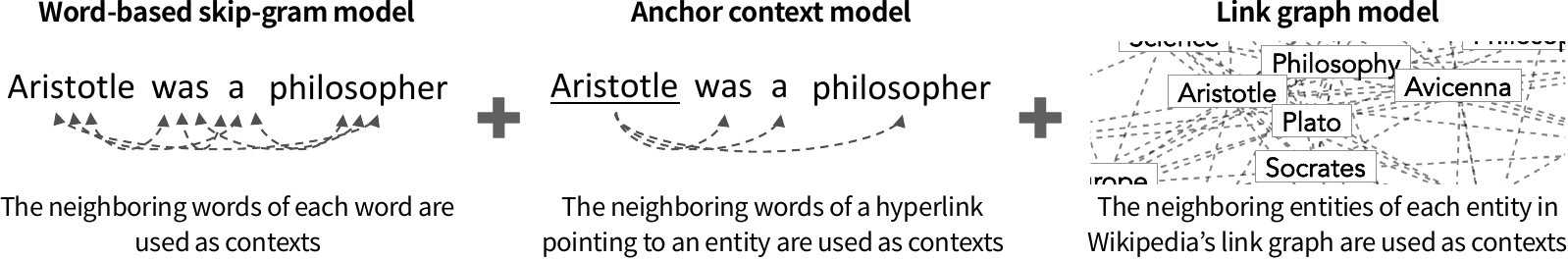}
    \caption{Wikipedia2Vec learns embeddings by jointly optimizing word-based skip-gram, anchor context, and link graph models.}
    \label{fig:model}
\end{figure*}

Wikipedia2Vec is an easy-to-use, optimized tool for learning embeddings from Wikipedia.
This tool can be installed using the Python's \texttt{pip} tool (\texttt{pip install wikipedia2vec}).
Embeddings can be learned easily by running the \texttt{wikipedia2vec train} command with a Wikipedia dump file\footnote{The dump file can be downloaded at Wikimedia Downloads: \url{https://dumps.wikimedia.org}} as an argument.
Figure \ref{fig:training} shows the shell commands that download the latest English Wikipedia dump file and run training of the embeddings based on this dump using the default hyper-parameters.\footnote{The \textit{train} command has many optional hyper-parameters that are described in detail in the documentation.}
Furthermore, users can easily use the learned embeddings.
Figure \ref{fig:api} shows the example Python code that loads the learned embedding file, and obtains the embeddings of an entity \textit{Scarlett Johansson} and a word \textit{tokyo}, as well as the most similar words and entities of an entity \textit{Python}.

\subsection{Model}
\label{subsec:model}

Wikipedia2Vec implements the conventional skip-gram model \cite{Mikolov2013,Mikolov2013a} and its extensions proposed in \newcite{Yamada2016} to map words and entities into the same $d$-dimensional vector space.
The skip-gram model is a neural network model with a training objective to find embeddings that are useful for predicting context items (i.e., words or entities in this paper) given each item.
The loss function of the model is defined as:
\begin{equation}
    \mathcal{L}_s = -\sum_{o_i \in O}\sum_{o_c \in C_{o_i}}\log P(o_c|o_i),
\end{equation}
where $O$ is a set of all items (i.e., words or entities), $C_o$ is the set of context items of $o$, and the conditional probability $\log P(o_c|o_i)$ is defined using the following softmax function:
\begin{equation}
    P(o_c|o_i) = \frac{\exp(\mathbf{V}_{o_i}\!^\top \mathbf{U}_{o_c})}{\sum_{o \in O}\exp(\mathbf{V}_{o_i}\!^\top \mathbf{U}_o)},
    \label{eq:skip-gram-softmax}
\end{equation}
where $\mathbf{V}_o \in \mathbb{R}^d$ and $\mathbf{U}_o \in \mathbb{R}^d$ denote the embeddings of item $o$ in embedding matrices $\mathbf{V}$ and $\mathbf{U}$, respectively.

Our tool learns the embeddings by jointly optimizing the three skip-gram-based sub-models described below (see also Figure \ref{fig:model}).
Note that the matrices $\mathbf{V}$ and $\mathbf{U}$ contain the embeddings of both words and entities.

\paragraph{Word-based Skip-gram Model}

Given each word in a Wikipedia page, this model learns word embeddings by predicting the neighboring words of the given word.
Formally, given a sequence of words $w_1, w_2, ..., w_N$, the loss function of this model is defined as follows:
\begin{equation}
    \mathcal{L}_w = -\sum_{i=1}^{N}\sum_{-c \leq j \leq c,j \neq 0}\log P(w_{i+j}|w_i),
\end{equation}
where $c$ is the size of the context words, and $P(w_{i+j}|w_i)$ is computed based on Eq.\eqref{eq:skip-gram-softmax}.

\paragraph{Anchor Context Model}

This model aims to place similar words and entities close to one another in the vector space using hyperlinks and their neighboring words in Wikipedia.
From a given Wikipedia page, the model extracts the referent entity and surrounding words (i.e., previous and next $c$ words) from each hyperlink in the page, and learns embeddings by predicting surrounding words given each entity.
Consequently, the loss function of this model is defined as follows:
\begin{equation}
    \mathcal{L}_a = -\sum_{(e_i, Q) \in A}\sum_{w_c \in Q}\log P(w_c|e_i),
\end{equation}
where $A$ denotes a set of all hyperlinks in Wikipedia, each containing a pair of a referent entity $e_i$ and a set of surrounding words $Q$, and $P(w_c|e_i)$ is computed based on Eq.\eqref{eq:skip-gram-softmax}.

\paragraph{Link Graph Model}

This model aims to learn entity embeddings by predicting the neighboring entities of each entity in the Wikipedia's link graph--an undirected graph whose nodes are entities and the edges represent the presence of hyperlinks between the entities.
We create an edge between a pair of entities if the page of one entity has a hyperlink to that of the other entity, or if both pages link to each other.
The loss function of this model is defined as:
\begin{equation}
    \mathcal{L}_e = -\sum_{e_i \in E}\sum_{e_o \in C_{e_i}}\log P(e_o|e_i),
\end{equation}
where $E$ is the set of all entities in the vocabulary, and $C_e$ is the neighboring entities of entity $e$ in the link graph, and $P(e_o|e_i)$ is computed by Eq.\eqref{eq:skip-gram-softmax}.

Finally, we define the loss function of our model by linearly combining the three loss functions described above:
\begin{equation}
    \mathcal{L} = \mathcal{L}_w + \mathcal{L}_a + \mathcal{L}_e
\end{equation}
The training is performed by minimizing this loss function using stochastic gradient descent.
We use negative sampling \cite{Mikolov2013a} to convert the softmax function (Eq.\eqref{eq:skip-gram-softmax}) into computationally feasible ones.
The resulting matrix $\mathbf{V}$ is used as the learned embeddings.

\subsection{Automatic Generation of Hyperlinks}
\label{subsec:link-generation}

Because Wikipedia instructs its contributors to create a hyperlink only at the first occurrence of the entity name on a page, many entity names do not appear as hyperlinks.
This is problematic for our anchor context model because it uses hyperlinks as a source to learn the embeddings.

To address this problem, our tool automatically generates hyperlinks using a mention-entity dictionary that maps entity names (e.g., ``apple'') to its possible referent entities (e.g., \textit{Apple Inc.} or \textit{Apple\_(food)}) (see Section \ref{sec:implementation} for details).
Our tool extracts all words and phrases from a Wikipedia page and converts each into a hyperlink to an entity if either the entity is referred to by a hyperlink on the same page, or there is only one referent entity associated with the name in the dictionary.

\section{Implementation}
\label{sec:implementation}

Our tool is implemented in Python and most of its code is compiled into C++ using Cython \cite{Behnel2011} to optimize the run-time performance.

As described in Section \ref{subsec:model}, our link graph and anchor context models are based on the hyperlinks in Wikipedia.
Because Wikipedia contains numerous hyperlinks, it is challenging to use them efficiently.
To address this, we introduce two optimized components--link graph matrix and mention-entity dictionary--that are used during training.

\paragraph{Link Graph Matrix}
During training, our link graph model needs to obtain numerous neighboring entities of an entity in a large link graph of Wikipedia.
To reduce latency, this component stores the entire graph in the memory using the binary sparse matrix in the compressed sparse row (CSR) format, in which its rows and columns represent entities and its values represent the presence of hyperlinks between corresponding entity pairs.
Because the size of this matrix is typically small, it can easily be stored on the memory.\footnote{The size of the matrix of English Wikipedia is less than 500 megabytes with our default hyper-parameter settings.}
Note that given a row index in the CSR matrix, the time complexity of obtaining its non-zero column indices (corresponding to the neighboring entities of the entity that corresponds to the row index) is $O(1)$.

\paragraph{Mention-entity Dictionary}
A mention-entity dictionary is used to generate hyperlinks described in Section \ref{subsec:link-generation}.
The dictionary maps entity names to their possible referent entities and is created based on the names and their referent entities obtained from all hyperlinks in Wikipedia.
Our tool extracts all words and phrases from a Wikipedia page that are included in the dictionary containing a large number of entity names.
To implement this in an efficient manner, we use the Aho--Corasick algorithm, which is an efficient string search algorithm using finite state machine constructed from all entity names.
After detecting the words and phrases in the dictionary, our tool converts them to hyperlinks based on heuristics described in Section \ref{subsec:link-generation}.

The embeddings are trained by simultaneously iterating over pages in Wikipedia and entities in the link graph in a random order.
The texts and hyperlinks in each page are extracted using the mwparserfromhell MediaWiki parser.\footnote{\url{https://github.com/earwig/mwparserfromhell}}
We do not use semi-structured data such as tables and infoboxes.
We also generate hyperlinks using the mention-entity dictionary.
We store the embeddings as a float matrix in a shared memory and update it using multiple processes.
Linear algebraic operations required to learn embeddings are implemented using C functions in Basic Linear Algebra Subprograms (BLAS).

Additionally, our tool uses a tokenizer to detect words from a Wikipedia page.
The following four tokenizers are currently implemented in our tool:
(1) the multi-lingual ICU tokenizer\footnote{\url{http://site.icu-project.org}} that implements the unicode text segmentation algorithm \cite{Davis2019}, (2) a simple rule-based tokenizer that splits the text using white space characters, (3) the Jieba tokenizer\footnote{\url{https://github.com/fxsjy/jieba}} for Chinese, and (4) the MeCab tokenizer\footnote{\url{https://taku910.github.io/mecab}} for Japanese and Korean.

\section{Experiments}

We conducted experiments to compare the quality and efficiency of our tool with those of the existing tools.
To evaluate the quality of the entity embeddings, we used the KORE entity relatedness dataset \cite{Hoffart2012}.
The dataset consists of 21 entities, and each entity has 20 related entities with scores assessed by humans.
Following past work, we reported the Spearman's rank correlation coefficient between the gold scores and the cosine similarity between the entity embeddings.
We used two popular entity embedding tools, RDF2Vec \cite{ristoski2018rdf2vec} and Wiki2vec, as baselines.

We also evaluated the quality of the word embeddings by employing two standard tasks:
(1) a word analogy task using the semantic subset (SEM) and syntactic subset (SYN) of the Google Word Analogy data
set \cite{Mikolov2013}, and
(2) a word similarity task using two standard datasets, namely SimLex-999 (SL) \cite{Hill:2015:SES:2893320.2893324} and WordSim-353 (WS) \cite{Finkelstein2002}.
Following past work, we reported the accuracy for the word analogy task, and the Spearman's rank correlation coefficient between the gold scores and the cosine similarity between the word embeddings for the word similarity task.
As baselines for these tasks, we used the skip-gram model \cite{Mikolov2013} implemented in the gensim library 3.6.0 \cite{rehurek_lrecb} and the extended skip-gram model implemented in the fastText tool 0.1.0 \cite{TACL999}.
We used WikiExtractor\footnote{\url{https://github.com/attardi/wikiextractor}} to create the training corpus for baselines.
To the extent possible, we used the same hyper-parameters to train our models and the baselines.\footnote{We used the following settings: $dim\_size=500$, $window=5$, $negative=5$, $iteration=5$}

We also reported the time required for training using our tool and the baseline word embedding tools.
Note that the training of RDF2Vec and Wiki2vec tools are implemented using gensim.

We conducted experiments using Python 3.6 and OpenBLAS 0.3.3 installed on the c5d.9xlarge instance with 36 CPU cores deployed on Amazon Web Services.
To train our models and the baseline word embedding models, we used the April 2018 version of the English Wikipedia dump.

\subsection{Results}

\begin{table}[tb]
    \centering
    \scalebox{0.76}{
        \begin{tabular}{lc}
            \hline
            Name                               & Score         \\
            \hline
            Ours                               & \textbf{0.71} \\
            Ours (w/o link graph model)        & 0.61          \\
            Ours (w/o hyperlink generation)    & 0.69          \\
            RDF2Vec \cite{ristoski2018rdf2vec} & 0.69          \\
            Wiki2vec                           & 0.52          \\
            \hline
        \end{tabular}
    }
    \caption{The results of Wikipedia2Vec and the baseline entity embeddings on the KORE dataset.}
    \label{tb:kore-results}
\end{table}

\begin{table}[tb]
    \centering
    \setlength{\tabcolsep}{2pt}
    \scalebox{0.76}{
        \begin{tabular}{lccccc}
            \hline
                                            & SEM           & SYN           & SL            & WS            & Time   \\
            \hline
            Ours                            & \textbf{0.79} & 0.68          & \textbf{0.40} & 0.71          & 276min \\
            Ours (w/o link graph model)     & 0.77          & 0.67          & 0.39          & 0.70          & 170min \\
            Ours (w/o hyperlink generation) & \textbf{0.79} & 0.67          & 0.39          & \textbf{0.72} & 211min \\
            Ours (word-based skip-gram)     & 0.75          & 0.67          & 0.36          & 0.70          & 154min \\
            gensim \cite{rehurek_lrecb}     & 0.75          & 0.67          & 0.37          & 0.70          & 197min \\
            fastText \cite{TACL999}         & 0.63          & \textbf{0.70} & 0.37          & 0.69          & 243min \\
            \hline
        \end{tabular}
    }
    \caption{The results of Wikipedia2Vec and the baseline word embeddings on the word analogy and word similarity datasets.}
    \label{tb:word-results}
\end{table}

Table \ref{tb:kore-results} shows the results of our models and the baseline entity embedding models of the KORE dataset.\footnote{We obtained the results of the RDF2Vec and Wiki2vec models from \newcite{ristoski2018rdf2vec}.}
\textit{w/o link graph model} and \textit{w/o hyperlink generation} are the results of ablation studies disabling the link graph model and automatic generation of hyperlinks, respectively.

Our model successfully outperformed the RDF2Vec and Wiki2vec models and achieved a state-of-the-art result on the KORE dataset.
The results also indicated that the link graph model and automatic generation of hyperlinks improved the performance of the KORE dataset.

Table \ref{tb:word-results} shows the results of our models with the baseline word embedding models on the word analogy and word similarity datasets.
We also tested the performace of the word-based skip-gram model implemented in our tool by disabling the link graph and anchor context models.

Our model performed better than the baseline word embedding models on the SEM dataset, as well as on both word similarity datasets.
This demonstrates that the semantic signals of entities provided by the link graph and anchor context models are beneficial for improving the quality of word embeddings.
Additionally, the feature of the automatic generation of hyperlinks did not generally contribute to the performance on these datasets.

Our implementation of the word-based skip-gram model was substantially faster than gensim and fastText.
Furthermore, the training time of our full model was comparable to that of the baseline word embedding models.

\section{Interactive Demonstration}

\begin{figure*}[tb]
    \centering
    \includegraphics[width=\textwidth,clip]{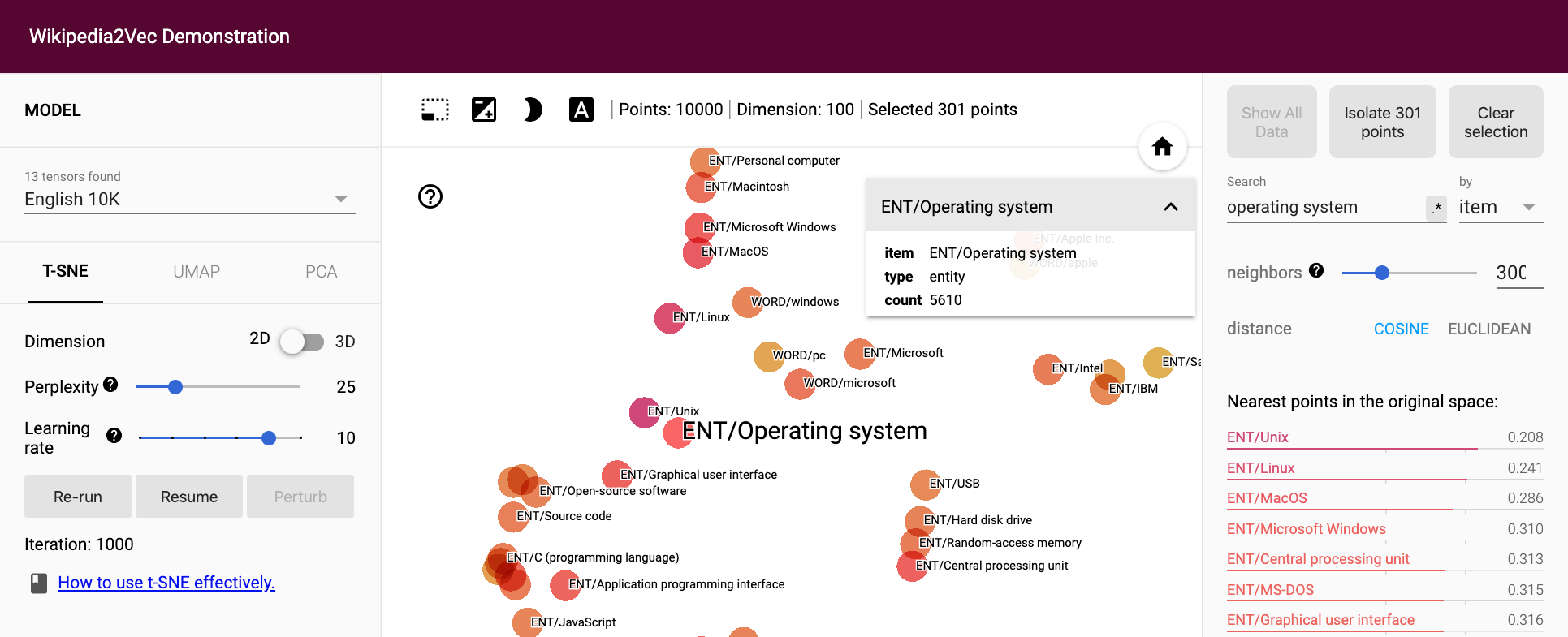}
    \caption{The screenshot of our web-based demonstration.
        Users can select the target embeddings (top left), configure the dimensionality reduction algorithm (bottom left), explore the visualized embeddings (center), and query similar words and entities based on an arbitrary word or an entity (right).}
    \label{fig:demo}
\end{figure*}

We developed a web-based interactive demonstration that enables users to explore the embeddings of words and entities learned by our proposed tool (see Figure \ref{fig:demo}).
This demonstration enables users to visualize the embeddings onto a two- or three-dimensional space using three dimensionality reduction algorithms, namely t-distributed stochastic neighbor embedding (t-SNE) \cite{maaten2008visualizing}, uniform manifold approximation and projection (UMAP) \cite{mcinnes2018umap}, and principal component analysis (PCA).
Users can move around the visualized embedding space by dragging and zooming using the mouse.
Moreover, the demonstration also allows users to explore the embeddings by querying similar items (words or entities) of an arbitrary item.

We used the pretrained embeddings of 12 languages released with this paper as the target embeddings.
Furthermore, we also provided the English embeddings trained without the link graph model to allow users to qualitatively investigate how the link graph model affects the resulting embeddings.

Our demonstration is developed by extending the TensorFlow Embedding Projector.\footnote{\url{https://projector.tensorflow.org}}
The demonstration is available at \url{https://wikipedia2vec.github.io/demo}.

\section{Use Cases}
\label{subsec:use-cases}
The embeddings learned using our proposed tool have already been used effectively in various recent studies.
\newcite{Poerner2019BERTQA} have recently demonstrated that by combining BERT with the entity embeddings trained by our tool outperforms BERT and knowledge-enhanced contextualized word embeddings (i.e., ERNIE \cite{Zhang2019ERNIE:Entities}) on unsupervised question answering and relation classification tasks, without any computationally expensive additional pretraining of BERT.
\newcite{10.1007/978-3-319-94042-7_10} developed a neural network-based question answering system based on our tool, and won a competition held by the NIPS 2017 conference.
\newcite{sato-EtAl:2017:I17-2}, \newcite{Chen2019YELM:Linking}, and \newcite{Yamada2019NeuralClassification} achieved state-of-the-art results on named entity recognition, entity linking, and text classification tasks, respectively, based on the embeddings learned by our tool.
Furthermore, \newcite{Papalampidi2019MovieIdentification} proposed a neural network model of analyzing the plot structure of movies using the entity embeddings learned by our tool.
Other examples include entity linking \cite{Yamada2016,eshel-EtAl:2017:CoNLL}, named entity recognition \cite{Lara-Clares2019}, paraphrase detection \cite{Duong2019ADetection}, fake news detection \cite{Singh2019OnArticles}, and knowledge graph completion \cite{Shah2019AnModels}.

\section{Conclusions}

In this paper, we present Wikipedia2Vec, an open-source tool for learning the embeddings of words and entities easily and efficiently from Wikipedia.
Our experiments demonstrate the superiority of the proposed tool in terms of the quality of the embeddings and the efficiency of the training compared to the existing tools.
Furthermore, our tool has been effectively used in many recent state-of-the-art models, which indicates the effectiveness of our tool on downstream tasks.
We also introduce a web-based interactive demonstration that enables users to explore the learned embeddings.
The source code and the pre-trained  embeddings for 12 languages are released with this paper.

\bibliography{references}
\bibliographystyle{acl_natbib}

\end{document}